\documentclass[letterpaper, 10 pt, conference]{ieeeconf}
\IEEEoverridecommandlockouts
\usepackage{cite}
\usepackage{amsmath,amssymb,amsfonts}
\usepackage{algorithmic}
\usepackage{graphicx}
\usepackage{gensymb}
\usepackage{textcomp}
\usepackage{xcolor}
\usepackage{siunitx}
\def\BibTeX{{\rm B\kern-.05em{\sc i\kern-.025em b}\kern-.08em
    T\kern-.1667em\lower.7ex\hbox{E}\kern-.125emX}}
\begin{document}

\title{\LARGE \bf Soft and Highly-Integrated Optical Fiber Bending Sensors for Proprioception in Multi-Material 3D Printed Fingers}

\author{Ellis Capp$^{1*}$, Marco Pontin$^{1}$, Peter Walters$^{2}$, and Perla Maiolino$^{1}$% <-this % stops a space
\thanks{This work was supported by  Engineering and Physical Sciences Research Council (EPSRC) Grant EP/V000748/1.}% <-this % stops a space
\thanks{$^{1}$Ellis Capp, Marco Pontin, and Perla Maiolino are with Oxford Robotics Institute, University of Oxford, UK;}%
\thanks{$^{2}$Peter Walters is  with the Department of Engineering Science, University of Oxford, UK.}%
\thanks{*Corresponding author: {\tt\small ellis@robots.ox.ac.uk}}}

\maketitle
\thispagestyle{empty}
\pagestyle{empty}

\begin{abstract}
Accurate shape sensing, only achievable through distributed proprioception, is a key requirement for closed-loop control of soft robots. Low-cost power efficient optoelectronic sensors manufactured from flexible materials represent a natural choice as they can cope with the large deformations of soft robots without loss of performance. However, existing integration approaches are cumbersome and require manual steps and complex assembly. We propose a semi-automated printing process where plastic optical fibers are embedded with readout electronics in 3D printed flexures. The fibers become locked in place and the readout electronics remain optically coupled to them while the flexures undergo large bending deformations, creating a repeatable, monolithically manufactured bending transducer with only 10 minutes required in total for the manual embedding steps. We demonstrate the process by manufacturing multi-material 3D printed fingers and extensively evaluating the performance of each proprioceptive joint. The sensors achieve 70\% linearity and 4.81° RMS error on average. Furthermore, the distributed architecture allows for maintaining an average fingertip position estimation accuracy of 12 mm in the presence of external static forces. To demonstrate the potential of the distributed sensor architecture in robotics applications, we build a data-driven model independent of actuation feedback to detect contact with objects in the environment. 
\end{abstract}

\section{Introduction}\label{sec:Introduction}
\begin{figure}
    \centering
    \includegraphics[width=0.9\columnwidth]{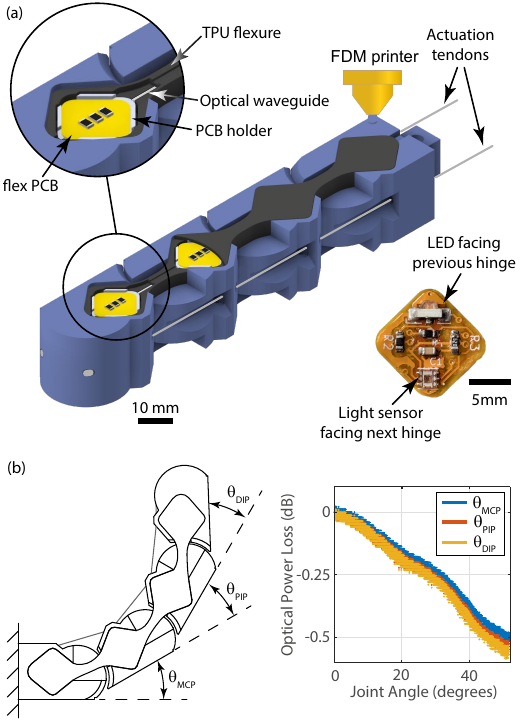}
    \caption{(a) The multi-material 3D printed finger alongside a photo of one of the embedded flex-PCBs. (b) Each joint in the TPU flexure holds an embedded optical fiber bending sensor enabling light loss-based pose estimation for each joint.}
    \label{fig:opening_figure}
    \vspace{-1.5em}
\end{figure}

Soft robots have attracted a wide interest among the robotics community. They can be used effectively in unstructured environments because of the inherent safety and adaptability afforded to them by their mechanical compliance. This makes them promising in applications such as human-robot interaction, manipulation, and exploration~\cite{polygerinos2017soft, shorthose2022design, becker2022active, brown2010universal, ng2023untethered, zhang2023progress}. However, while progress has been made in these domains, soft robots are still limited by their sensing capabilities as their virtually infinite degrees of freedom require sensors to be distributed throughout the robot body~\cite{toward_perceptive_soft_robots_wang}. This creates a two-fold challenge: sensors must be sufficiently flexible to survive the high strains experienced by soft robots while maintaining good performance, and they need to be easily integrable at the manufacturing stage~\cite{flexible_sensing_qu}.

To tackle this challenge, researchers have developed a host of sensors of wide-ranging nature, such as resistive, capacitive, and optics-based ones~\cite{Truby2019,Toshimitsu2021,Georgopoulou2023,optoelectronically_innervated_hand_zhao}. Optical sensors in particular show potential because of their low hysteresis, high sensitivity, high strain performance, and their insensitivity to electromagnetic interference and temperature~\cite{flexible_sensing_qu,Hegde2023}. However well these sensors may perform, their fabrication and integration into soft robots remains cumbersome, especially when it comes to the necessary electronics. For example, Sareh et al. attach their macrobend stretch sensor onto the outside of a soft arm by sewing plastic optical fibers (POFs) in a looping pattern to the strain-limiting braided sleeve~\cite{macrobend_optical_sensing_sareh}, and Yang et al. externally route POFs to each of the sensing locations in their soft robotic finger~\cite{Yang2020}. Both approaches achieve sensing of the multidimensional state of a soft robot, but are unable to scale to distributed sensing or provide an untethered solution without external optical wiring because they do not integrate the readout electronics, hindering robot movement and durability~\cite{Rich2018}. 
\begin{figure*}[ht]
    \centering
    \includegraphics[width=0.9\textwidth]{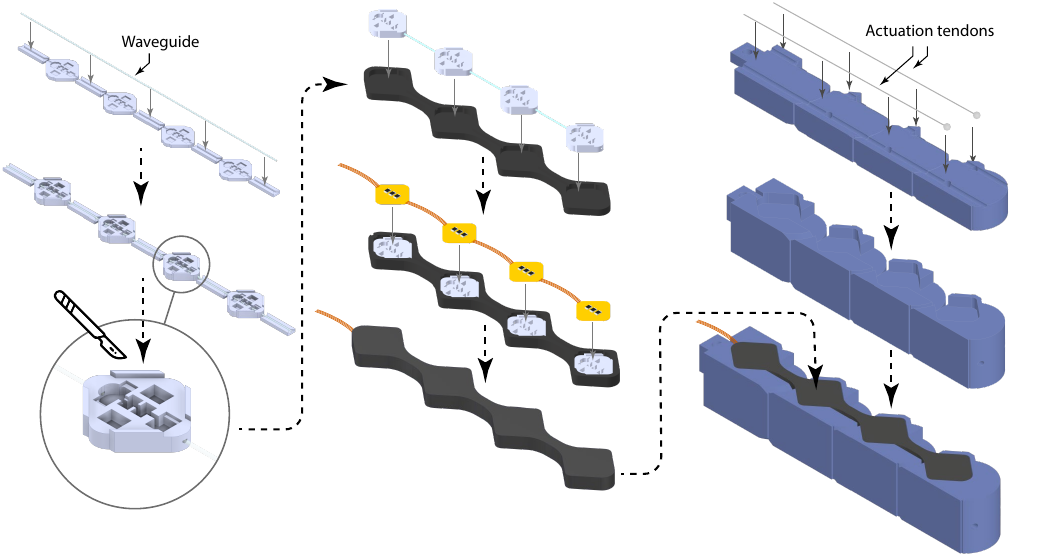}
    \caption{Schematic representation of the manufacturing and integration process of the sensorized multi-material finger. There are three main steps to the process: embedding of the optical fiber, embedding of the flex-PCBs, and printing of the finger. All the embedding steps are conducted during the printing process by temporarily pausing filament extrusion.}
    \label{fig:fabrication_figure}
    \vspace{-1em}
\end{figure*}

An alternative method that achieves a higher degree of integration is to use the optical waveguides as part of the structure of the soft robot, either by direct fabrication of soft optical waveguides into the robot body~\cite{Yun2021,Jung2020,DelBono2024} or using the waveguides as a structural component~\cite{optoelectronically_innervated_hand_zhao,Kang2023}. Direct fabrication approaches usually entail complicated multi-step mechanical processes including molding and casting, laser micromachining, or layer deposition of expensive reflective materials. In contrast, using the optical waveguides as a structural component leads to simpler integration. This idea lends itself particularly well to tendon-driven robotic fingers, where the actuation tendon itself can be made from an optical waveguide. In \cite{Yi2023}, Yi et al. implement force sensing employing a fiber Bragg grating (FBG) in series with the tendon, while in \cite{Han2024} Han et al. train a neural network to predict the pose of a finger from the optical power loss in the finger's soft optical tendon. Despite both of these robotic fingers achieving a design that elegantly integrates one sensor into their structure, they lack the distributed sensing required to disentangle proprioception and exteroception: the FBG tendon is only used for exteroception, whilst the soft optical tendon is only able to predict poses if there are no external forces. Moreover, despite making use of 3D printing techniques, neither approach achieves monolithic integration at the manufacturing step, which remains an exciting and very recent development in additive manufacturing for sensorized soft robots~\cite{Lipson2015,Rus2015,Muth2014,Shih2019,Ntagios2020}.

This study aims to apply fused filament fabrication (FFF) 3D printing, a cheap and widely accessible technology, to implement an optical sensing solution that is both distributed and highly integrated in order to achieve proprioception in the presence of external disturbances. We contribute a method of reliably embedding POFs into a multi-material 3D printed finger with optical bending sensors embedded in each compliant joint (Fig. \ref{fig:opening_figure}). To achieve distributed sensing, we also completely embed the LEDs and readout electronics with the POFs to produce a monolithically manufactured sensorized flexure, contributing to the largely unexplored area of embedding circuitry into soft robots~\cite{Woodman2024}. These electronics cost a total of \$10 per finger with a low power consumption of \qty{188}{mW}, providing scalability especially when compared to FBG-based sensors. By embedding during the 3D printing stage, we eliminate both the need for manual assembly of parts after printing and the presence of any external wiring or optics, resulting in a tendon-actuated finger with fully self-contained sensing. This research represents a step towards the integrated, distributed sensing necessary for fully autonomous soft robots \cite{Soter2018,Thuruthel2019,Truby2020}. 

The remainder of this paper is structured as follows: in section \ref{sec:materials_and_methods}, we provide details of the embedding method and multi-material finger design. In section \ref{sec:experimental_calibration}, we describe our sensor characterization procedure and results. In section \ref{sec:experimental_validation}, we describe two validation experiments and discuss their results. In the first experiment, we compare the sensors' ability to predict fingertip position in the presence of varying static forces compared to a baseline finger with only one optical bending sensor along its length. In the second experiment, we train a model on sensor data in the absence of external contact and use it to predict contact. We provide our final remarks in section \ref{sec:conclusion}. 

\section{Materials and Methods}\label{sec:materials_and_methods}
The creation of the multi-material sensorized finger requires a short sequence of mostly automated manufacturing steps, displayed in Fig.  \ref{fig:fabrication_figure}. In this section, we analyze these steps in detail. In particular, the manual embedding step in which the POFs are encased in holders and the subsequent manual embedding steps in which the POFs, holders, and FPCBs are embedded into the flexure take 10 minutes in total and requires no adhesives and no special tools other than a razor blade and a soldering iron. 
\subsection{Optical Fiber Embedding Step}
\begin{figure}
    \centering
    \includegraphics[width=0.9\columnwidth]{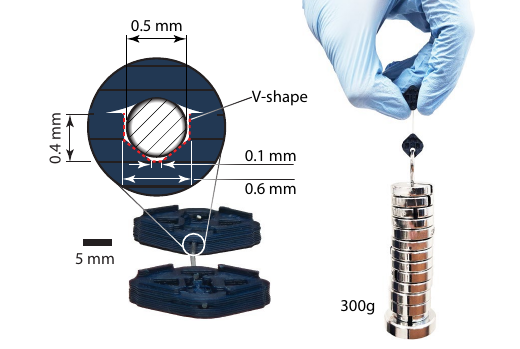}
    \caption{Closeup photo of a pair of holders with an embedded optical fiber and a schematic detailed view of the embedding channel. The embedded fiber is able to support a mass of \qty{300}{\g}.}
    \label{fig:clearances_and_embed_closeups}
    \vspace{-1em}
\end{figure}

The fabrication process started with the embedding of the POFs. For this, we implemented a pause-and-place method where the the 3D printing of custom-designed holders with channels to accommodate the fiber was temporarily halted to enable the placement of the POFs before printing was resumed. The pause was scheduled at the layer where the channel was completed, to ensure good localization of the POF and minimize damage due to the interaction with the 3D printer nozzle.

Unjacketed ESKA POF was used with a diameter of \qty{500}{\um}, a commonly available POF manufactured from polymethyl methacrylate. The holders were printed on a Tenlog TL-D3 Pro independent dual extruder (IDEX) 3D printer (Innocube3D, Shenzhen, China) out of polylactic acid (PLA), with extruder temperature set to \qty{210}{\degree C}, a layer height of \qty{0.2}{\mm}, and a print speed of \qty{60}{\mm/s}. In preliminary experiments, we utilized POFs with a black polyethylene jacket with outer diameter \qty{1}{\mm} and core diameters of \qty{250}{\um} and \qty{500}{\um}. However, we found that when attempting to embed these POFs, the polyethylene jacket was damaged by the extruder nozzle during the printing process and allowed the fibers to easily slip out. We therefore switched to unjacketed POFs with core diameters of \qty{1000}{\um}, \qty{500}{\um}, and \qty{250}{\um}. We chose \qty{500}{\um} POFs to manufacture the sensors in this study because the \qty{1000}{\um} POFs exhibited high plastic strain at the bending radii required for the robotic finger geometry, while the \qty{250}{\um} POFs were challenging to embed repeatably due to the tolerances of our 3D printer.

When readying the holders for printing, we used the slicing software Cura to place them at the same spacing as the final spacing of the finger phalanges, \qty{28}{\mm} apart. One holder was printed for each of the four phalanges and a single fiber was embedded running through all four holders. After printing, the POF was cut with a surgical blade, as required to complete the POF-holder assembly. 

To enhance the repeatability and holding strength of the fiber embedding, we designed the fiber-accommodating channel with a V-shape as shown in Fig. \ref{fig:clearances_and_embed_closeups} to help the fiber self-center when pushed downward by the extruder nozzle. The height of the channel was \qty{0.4}{\mm}, equal to two layer heights, with a width of \qty{0.6}{\mm} on the higher layer to allow the fiber to be set into place and a width of \qty{0.1}{\mm} on the lower layer to form the V-shape. With this design, the fiber sits proud of the holders when the print is resumed so that the next extruded layer presses tightly against the top of the fiber. To further improve printing success, sacrificial jigs with channels to hold the fiber were printed in between each holder and the wall printing order was modified in Cura so that the printer would lock the POFs in place first before moving on to printing the remainder of the holders. A \qty{15}{\mm}$\times$\qty{15}{\mm}$\times$\qty{1.6}{\mm} cuboid was printed alongside the holders and jigs to prevent under-extrusion of the post-pause printed layer, and the printing temperature was temporarily increased to \qty{230}{\degree C} to promote inter-layer adhesion. The embedding process just described resulted in a reliable axial locking of the fibers, which could withstand loads of \qty{300}{g} without failing (Fig. \ref{fig:clearances_and_embed_closeups}).

\subsection{Flexible Printed Circuit Boards and Holder Design}
The FPCBs are based on the OPT3002 light-to-digital converter (LDC) (Texas Instruments, Dallas, TX, USA) and have a \qty{10}{\mm}$\times$\qty{10}{\mm} outline. We placed the LDC, supporting components, and a side-view 3010 LED on one side of each FPCB. We used the other side of the FPCBs for a selectable jumper to set the address of each LDC on the inter-integrated circuit (I2C) bus for power and communication wires.   

The highest spectral response of the LDC occurs at a wavelength of \qty{505}{nm}, so a yellow LED with a peak wavelength of \qty{590}{nm} was used. Because this response is in the visible spectrum, the LDC is sensitive to ambient light. We accepted this compromise because of the LDC's digital readout and miniaturized off-the-shelf package and we mitigated the effects of ambient light by 3D printing the outer casing of the sensor in black. 

The holders that lock the POFs in place also hold the FPCBs. The holders are designed with cavities that fit the circuit components of the FPCB and a \qty{0.6}{\mm} lip around the perimeter of the FPCB. We chose this height for the lip because it was the tallest height that could be achieved without interfering with the 3D printer nozzle during the sensor embedding process.

\subsection{Sensor Embedding Steps}
To prepare the FPCBs for embedding, we utilized a 3D printed jig to hold the FPCBs at the phalanx spacing before soldering \qty{0.15}{\mm} diameter enameled copper wires between each FPCB for the power and communication bus. After soldering, we encapsulated the solder joints to the wires in superglue to provide strain relief. We then embedded both the POF-holder and FPCB assemblies in the 3D printed thermoplastic polyurethane (TPU) (Filaflex 82A, Recreus, Elda, Spain) flexure in two pausing steps. During the first pause, we placed the POF-holder assembly into a cavity in the flexure. The printer then encapsulated the POFs while leaving the holders exposed to accept the FPCBs. During the second pause, the FPCB assembly was placed into the holders and a soldering iron with a chisel tip at \qty{350}{\degree C} was used to partially melt the midpoints of the lips over the sides of the FPCBs to lock them in place before resuming the print again. 

\subsection{Multi-material Finger Design}\label{subsec:finger_design}
We designed the multi-material finger with rigid phalanges printed from PLA and soft flexures connecting the phalanges printed from soft TPU with shore hardness 82A. The soft TPU gives the finger joint compliance and eliminates the need for any assembly or fasteners in the design, while the use of rigid PLA controls the kinematics of the actuated finger to allow the sensor angle readouts to predict the finger's shape. To facilitate printing with the TPU filament, we replaced one of the TL-D3 Pro's extruders with a Micro Swiss NG extruder (Ramsey, Minnesota, USA) and used print settings following the manufacturer's recommendations. 

The joints consist of a sliding contact pair between an inner and outer circle with \qty{0.12}{\mm} clearance in the undeformed state with the TPU flexures arranged symmetrically on both sides of the finger. The sliding contact pair geometry was selected as it provided a linear optical power loss. The sensors are unable to sense bending direction as their transduction principle is based on optical waveguide macrobending loss, therefore we included mechanical stops in the joint design to restrict bending to one direction. 

The finger is tendon-actuated, with the tendons made from \qty{0.4}{\mm} nylon fishing line and embedded at the mid-plane of the finger using another pause-and-place procedure. We secured the tendons to the distal phalanx by attaching split shot fishing weights to the tendons, placing the fishing weights in printed voids, and filling the voids around the fishing weights with superglue. 

The sensors are only embedded in the flexures on one side of the finger. To facilitate sensor characterization, flexures containing sensors were printed separately and inserted into a shaped cavity on the multi-material finger, held in place by compression. TPU flexures that did not contain sensors were printed with a high infill of 80\%. 

\section{Experimental Sensor Calibration}\label{sec:experimental_calibration}\begin{figure}
    \centering
    \includegraphics[width=0.9\columnwidth]{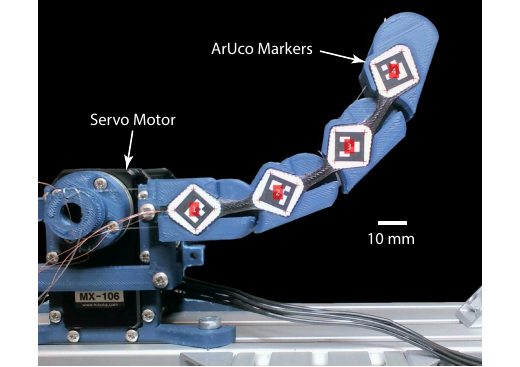}
    \caption{Setup used for the characterization of the fingers. The photo is taken according to the POV of the camera used for the tracking of the markers.}
    \label{fig:setup_figure}
    \vspace{-1em}
\end{figure}
\begin{figure*}[t]
\centering
\includegraphics[width=0.9\textwidth]
{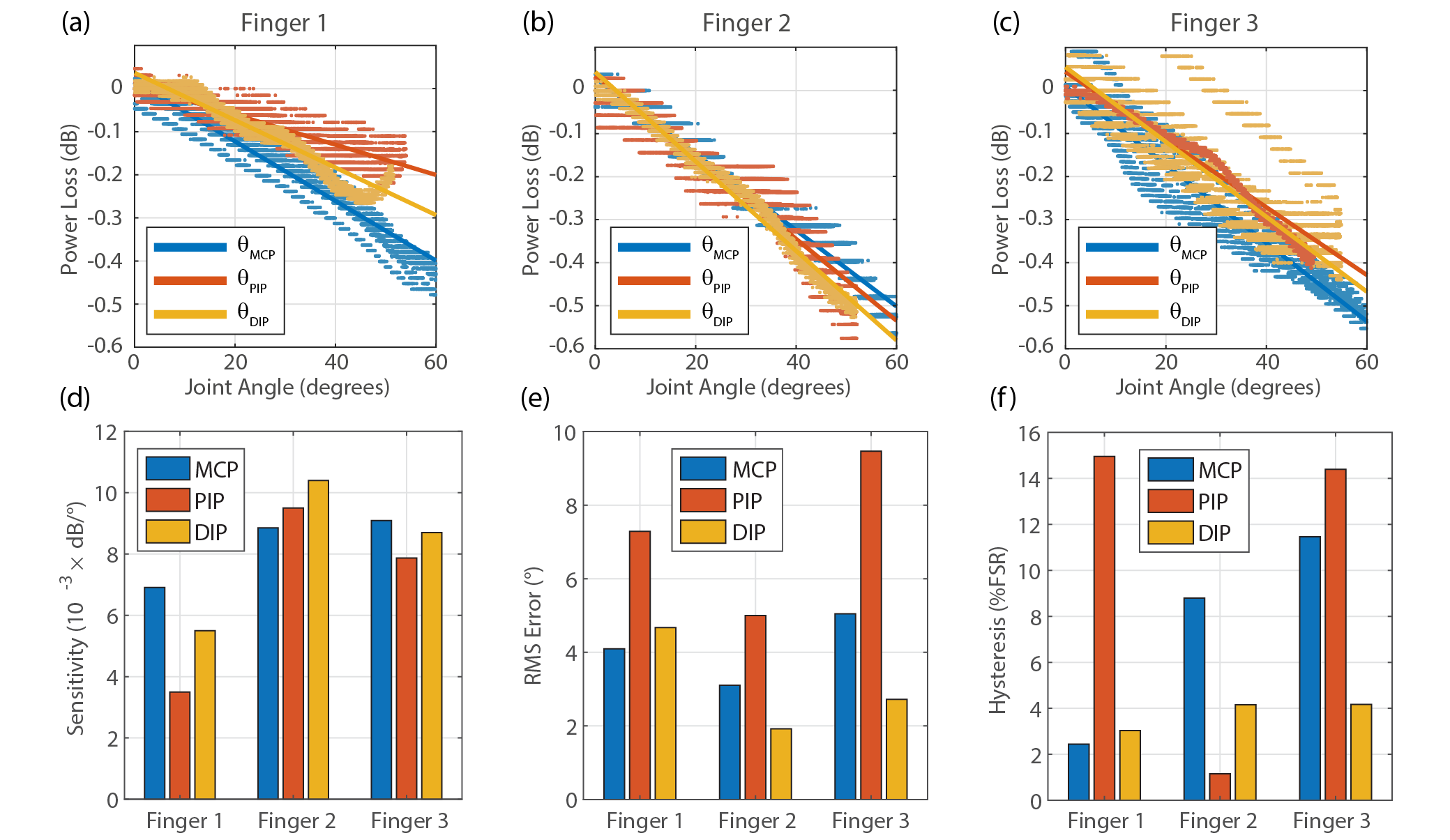}
\caption{Characterizations of the sensors. (a),(b),(c) The plots show measured optical power loss against the MCP, PIP, and DIP joint rotations for each finger during the quasi-static trials with their respective best-fit lines. (d) The sensitivity achieved for each sensor is shown grouped by finger on the bar chart. (e) RMS error in degrees for each joint is shown. (f) Hysteresis as percentage of full scale range for each joint is shown.}
\label{fig:fingers_results}
\vspace{-1em}
\end{figure*}
In this section, we describe a series of tests in which we analyze characteristics of the sensors that are important for engineering and control applications. Specifically, we characterize the reproducibility, repeatability, sensitivity, linearity, and hysteresis of the sensors, while also investigating the sensors' ability to retain their response characteristics over their operational life. Three sensorized flexures were 3D printed to evaluate sensor fabrication reproducibility, each of which were inserted into the same multi-material finger body.
\subsection{Protocol}\label{subsec:protocol}
\subsubsection{Quasi-Static Test}
To evaluate the sensor performance as-actuated, we used a 2-step pulley mounted on a Dynamixel MX-106R servo motor (Robotis, Seoul, South Korea) to actuate both the flexor and extensor tendons of the multi-material finger. After applying a \qty{15}{rpm} \qty{90}{\degree} transient rotation to the pulley, we observed a settling time under \qty{125}{\ms}. To characterize the sensors in quasi-static conditions, the pulley was rotated from \qty{0}{\degree} to \qty{263.6}{\degree} in increments of \qty{0.44}{\degree}. After each rotation, a pause of an eight of a second was performed before taking a measurement from the sensors. The entire procedure was repeated five times for each of three different fingers, resulting in $N=12010$ readings from each joint of each finger over the full test. The test was conducted with the apparatus placed under an LED lamp to control for ambient light. 

A webcam was placed square to the finger to enable joint angle measurements via image processing by detecting the pose of ArUco markers~\cite{ArUco} affixed to each phalanx of the finger (Fig. \ref{fig:setup_figure}) and taking the Euler angle of the markers in the plane of the webcam image. The rotations $\theta_\textrm{MCP}$, $\theta_\textrm{PIP}$, and $\theta_\textrm{DIP}$ of the metacarpophalangeal, proximal interphalangeal, and distal interphalangeal joints were then calculated by the difference between the marker rotations of their respective phalanges. The servo motor, microcontroller, and webcam were all controlled by the same laptop running a data collection script in MATLAB. 

\subsubsection{Stress Test}
To evaluate the sensor's ability to retain its response characteristics over time, we selected one finger and subjected it to cyclic motion over the full \qty{263.6}{\degree} motion range of the setup. We chose an angular speed for the tendon pulley of \qty{60}{rpm} corresponding with an actuation frequency of \qty{1.46}{Hz}, in line with existing actuation mechanisms in soft robots \cite{Li2023}. After performing all other experiments and applying 50 cycles to the finger to break it in, we applied 450 cycles in constant lighting conditions. Sensor data was acquired at \qty{8}{Hz}.

\subsection{Sensor Model}\label{subsec:sensor_model}
To obtain a calibration line for each sensor, we applied a least-squares linear regression to the data collected from all five quasi-static trials for each sensor according to a linear model of the form 
\begin{equation}\label{eq:model}
   \hat{y}=\begin{bmatrix}1 & \theta\end{bmatrix}\begin{bmatrix}\beta_0 \\ \beta_1\end{bmatrix}
\end{equation}
where $\hat{y}$ is the best fit of the sensor output $y$, $\begin{bmatrix}\beta_0 & \beta_1\end{bmatrix}^\top$ is the vector of regression coefficients. $y$ is expressed as optical power loss in decibels and is calculated as $y=-10\log_{10}(I/I_0)$, where $I$ is the raw digital value in nW/cm$^{2}$ provided by the sensor and $I_0$ is the raw value measured at the beginning of each trial.

\subsection{Results}
To evaluate the utility of the sensors for closed-loop control of soft robots, we examine the quality of the models in predicting the joint angle by calculating metrics for reproducibility, repeatability, sensitivity, linearity, and hysteresis. We first examined reproducibility by conducting an ANCOVA test using $\theta$ as the covariate and individual sensor responses as categorical factors. We observed a highly statistically significant difference between the responses [F(8,108072)=9822, p\textless 0.01], indicating that unique calibration lines should be used for each sensor. 

The sensitivity of each sensor was taken to be the slope of the calibration line, $\beta_1$. Percentage linearity was calculated as a percentage of full scale range (FSR) according to the formula in \cite{Fleming2013}: \begin{equation}
    \textrm{Linearity(\% FSR)}=\left(1-\frac{\max|y-\hat{y}|}{\max |y|}\right)\times 100\%
\end{equation}
To give a metric for repeatability, we calculated the RMS error measured in degrees $\sigma_\textrm{RMS}$ of each sensor according to the formula:
\begin{equation}
    \sigma_\textrm{RMS}=\frac{1}{\beta_1}\sqrt{\sum_{i=1}^N\frac{\left(y_i-\hat{y}_i\right)^2}{N-1}}
\end{equation}
The linearity values we achieved typically ranged between 59\% (PIP joint of finger 1) and 88\% (DIP joint of finger 2) with a mean value of 70\% and a single outlier due to the PIP joint of finger 3 with a value of 37\%. We note from these calibrations that the sensitivities are higher for fingers 2 and 3 than for finger 1. When taken with the results of the ANCOVA test, this suggests that the manual steps in the fabrication process, although limited, still affect the resulting sensor array.

To quantify hysteresis for each sensor, we took the difference between the mean sensor reading from the loading portions of the quasi-static trials and the mean sensor reading from the unloading portions of the quasi-static trials. We express this difference as a percentage of the FSR for each sensor in Fig. \ref{fig:fingers_results}f. We note that sensors with greater hysteresis also show less repeatability, as expected. 

\begin{figure}
\centering
\includegraphics[width=0.9\columnwidth]{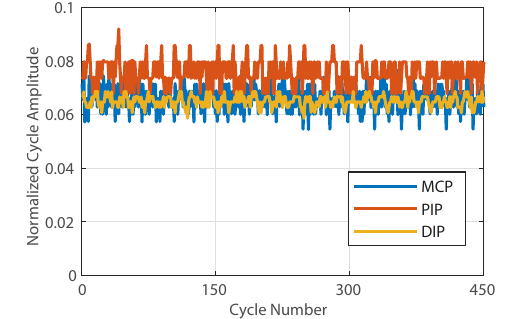}
\caption{Results of a stress test on one of the fingers. All sensors survived undergoing 450 actuation cycles, as demonstrated by their normalized cycle amplitude that do not decay over time.}
\label{fig:stress}
\vspace{-1em}
\end{figure}

We quantified the results of the stress test by measuring the normalized amplitude $(I_\textrm{max}-I_\textrm{min})/I_\textrm{max}$ during each cycle. Finger 3 completed the test with no detectable change in cycle amplitude on any of the three sensor readings (Fig. \ref{fig:stress}). This result indicates that the sensor is able to retain its response characteristics over a typical research service life. 

\section{Experimental Validation}\label{sec:experimental_validation}In this section, we describe the protocol, models used, and results of two experiments intended to validate that our distributed proprioceptive sensor is able to detect and reject external disturbances. In the first experiment, we applied static forces to the fingertip using calibration weights and examined the fingertip position error based on linear regression trained during the experiments of section \ref{sec:experimental_calibration}, and we compared this to a baseline single-sensor finger consisting of a single POF sensor integrated using the same 3D printing method as the multi-sensor fingers. In the second experiment, we trained a data-driven model using the data acquired during the experiments of section \ref{sec:experimental_calibration} to detect between the finger and a light switch based on deviations of the finger sensor readings from those acquired during actuation in the absence of contact. 
\subsection{Protocol}
\subsubsection{External Disturbance Experiment}
\begin{figure}
\centering
\includegraphics[width=0.9\columnwidth]{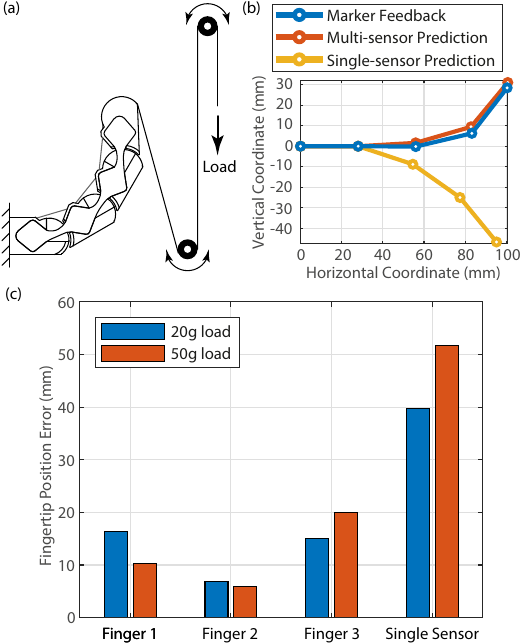}
\caption{(a) Diagram of experimental setup for applying static loads to fingertip. (b) Pose estimations of the highly integrated finger and single-sensor finger shown with marker feedback. The optical intensity measured by the single-sensor finger increases due to the applied force, resulting in a pose estimation pointing the opposite direction, inconsistent with the operating range of the sensor. (c) Average fingertip position error recorded during each trial.}
\label{fig:disturbances}
\vspace{-1em}
\end{figure}

To test the accuracy of our multi-sensor approach in the presence of external disturbances, we applied the same testing protocol as in \ref{subsec:protocol} to each finger with calibration weights tied to the fingertip with nylon fishing line (Fig. \ref{fig:disturbances}a). We ran one test with a \qty{20}{g} weight and one test with a \qty{50}{g} weight for each finger. We then compared the results with those coming from a finger of the same geometry with only one fiber optic running along its length and a single pair of FPCs, one emitter and one receiver. We first tested this version of the finger without weights to determine the baseline calibration for pose estimation purposes and then with the \qty{20}{g} and \qty{50}{g} weights, following the same protocol as before. 

\subsubsection{Contact Detection Experiment}
To demonstrate how the finger's multi-sensor architecture can be leveraged to detect contact with an object in the environment without any knowledge of the actuation state, we selected one finger, placed it underneath a light switch, and actuated it until it flipped the switch while collecting data from the sensors at \qty{8}{Hz}. We actuated the finger in steps of \qty{9}{\degree} with \qty{0.25}{\s} pauses to facilitate webcam image acquisition.

\begin{figure*}
    \centering
    \includegraphics[width=0.8\textwidth]{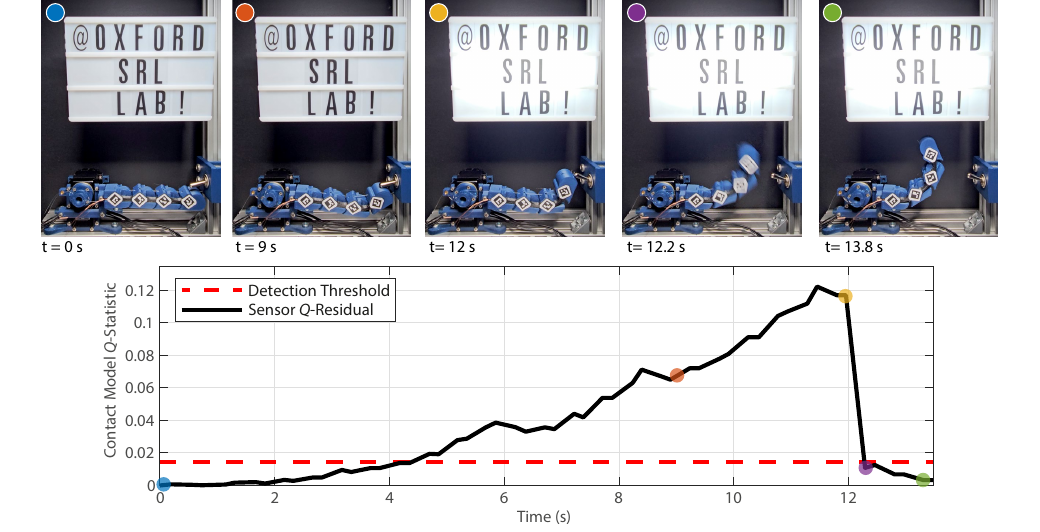}
    \caption{Contact detection experiment. The $Q$-statistic from the PCA model of the sensors is used against a threshold value shown in red in the chart to detect external contacts.}
    \label{fig:contact_detection_figure}
    \vspace{-1em}
\end{figure*}

\subsection{Pose Estimation and Contact Detection Models}
For pose estimation, we modeled the phalanges as a kinematic chain in two dimensional space with four rigid links of length \qty{28}{\mm} connected by joints with angles estimated by the sensors. For the multi-sensor fingers, joint angles were estimated by inverting the models obtained in section \ref{subsec:sensor_model}. For the single-sensor finger, we employed the same linear regression approach as the multi-sensor finger to train a model predicting each of the joint angles using the data from the single sensor acquired during the test with no weight.

For the contact detection model, we applied principle component analysis (PCA) to the sensor readings acquired from the quasi-static tests utilizing a technique commonly used for fault detection in industrial processes \cite{Yin2014}. We trained an $m$-dimensional PCA model with $m=3$ on the three-dimensional sensor data from the finger. The first component of the PCA model explains 91\% of the data variability, which is consistent with the one-DOF actuation of the finger and allowed us to conclude that the principal subspace of the PCA model has dimensionality $\beta=1$. The model detects contact by projecting online sensor observations onto the residual subspace of the PCA model and calculating the $Q$-statistic:
\begin{equation}\label{eq:SPE}
    Q=z^\top P_\textrm{res} P_\textrm{res}^\top z
\end{equation}
where $z$ is the $3\times 1$ vector of online sensor observations and $P_\textrm{res}$ is the matrix of residual PCA coefficients. We used the threshold for anomaly detection from \cite{Yin2014} determined with confidence level $\alpha=\text{90\%}$:
\begin{equation}\label{eq:J_th}
    J_{{\rm th,} Q} = \vartheta_{1}\left(\frac{c_{\alpha}\sqrt{2\vartheta_{2}h_{0}^{2}}}{\vartheta_{1}} + 1 + \frac{\vartheta_{2}h_{0}(h_{0} - 1)}{ \vartheta_{1}^{2}}\right)^{1/ h_{0}}
\end{equation}
where $c_\alpha=1.282$ is the normal deviate for the upper $1-\alpha$ percentile and $h_0$ and $\vartheta_i$ are parameters calculated from the variances of the last $m-\beta$ principal components $\lambda_i$:
$$h_{0} = 1 - \frac{2\vartheta_{1}\vartheta_{3}}{3\vartheta_{2}^{2}} \qquad \vartheta_{i} = \sum_{j = \beta + 1}^{m}(\lambda_{j})^{i}; \quad i = 1, 2, 3.$$

\subsection{Results}
The results from the external disturbance experiment (Fig. \ref{fig:disturbances}c) show that the multi-sensor finger estimates the fingertip position with lower error than the single-sensor finger. We expected based on the underactuated training data that the single-sensor prediction would produce approximately equal estimates for each joint angle and that the primary source of pose estimation error would be due to poses with unequal angles. However, we also see in Fig. \ref{fig:disturbances}b that the single-sensor pose prediction points in the wrong direction. This is due to the fact that the optical intensity reading of the sensor increases in the presence of external forces to the point that the sensors observe optical intensity gain rather than loss during the experiment. Because the sensor model is based on a linear regression that assumes the sensor will only experience optical intensity loss, if gain is observed, the sensor will predict a negative angle outside of the designed operational range. While this effect is also present in the multi-sensor fingers, it is mitigated by the redundancy afforded by the multi-sensor architecture. We also note that the magnitude of the weight did not directly correlate to the prediction error across the fingers, at least for the range of weights that was tested. 

The results from the contact detection experiment are displayed in Fig. \ref{fig:contact_detection_figure} and the supporting video to the paper. First contact was observed in the trial video at $t=\qty{0.7}{s}$, and the model detected contact at $t=\qty{4.2}{s}$. At $t=\qty{12.2}{s}$, the finger came out of contact with the light switch, and the corresponding $Q$ dropped below $J_{{\rm th,} Q}$ shortly afterwards, at the next sensor reading update. We note that the finger is able to detect coming out of contact online at \qty{8}{Hz}, while the webcam acquisition misses a \qty{0.25}{\s} frame due to motion blur then takes another \qty{0.25}{\s} frame to detect the new pose of the finger, further demonstrating the merits of a low processing-power distributed device. 

\section{Conclusion}\label{sec:conclusion}
During this study, we accomplished our aim of using FFF 3D printing to fabricate an optical sensor that can be both distributed and highly integrated and we demonstrated how this fabrication technique could be used to achieve proprioception in a multi-material 3D printed finger. Our compliant tendon-actuated finger is able to perform pose estimation in the presence of static forces and, with training data, it is able to detect contact with the environment.

There are still aspects that warrant further improvement and investigation. In particular, further development of the embedding process is needed to increase its reproducibility and improve overall output repeatability among different sensors. In addition, future work needs to build on our contribution by focusing on how the embodiment of the sensor affects its response characteristics. Our experience and the results from the study suggest that better sensor output might be achieved by optimizing the flexure morphology. Thanks to the accessibility and design flexibility of FFF 3D printing and the low-cost power efficient nature of our proposed sensing methodology, we envision it being adapted to existing soft robotics design and used in larger distributed sensing networks to finally achieve accurate proprioception in the soft robots of the future.

\bibliographystyle{IEEEtran}
\bibliography{references.bib}

\begin{thebibliography}{10}
\providecommand{\url}[1]{#1}
\csname url@samestyle\endcsname
\providecommand{\newblock}{\relax}
\providecommand{\bibinfo}[2]{#2}
\providecommand{\BIBentrySTDinterwordspacing}{\spaceskip=0pt\relax}
\providecommand{\BIBentryALTinterwordstretchfactor}{4}
\providecommand{\BIBentryALTinterwordspacing}{\spaceskip=\fontdimen2\font plus
\BIBentryALTinterwordstretchfactor\fontdimen3\font minus
  \fontdimen4\font\relax}
\providecommand{\BIBforeignlanguage}[2]{{%
\expandafter\ifx\csname l@#1\endcsname\relax
\typeout{** WARNING: IEEEtran.bst: No hyphenation pattern has been}%
\typeout{** loaded for the language `#1'. Using the pattern for}%
\typeout{** the default language instead.}%
\else
\language=\csname l@#1\endcsname
\fi
#2}}
\providecommand{\BIBdecl}{\relax}
\BIBdecl

\bibitem{polygerinos2017soft}
P.~Polygerinos, N.~Correll, S.~A. Morin, B.~Mosadegh, C.~D. Onal, K.~Petersen,
  M.~Cianchetti, M.~T. Tolley, and R.~F. Shepherd, ``Soft robotics: Review of
  fluid-driven intrinsically soft devices; manufacturing, sensing, control, and
  applications in human-robot interaction,'' \emph{Advanced engineering
  materials}, vol.~19, no.~12, p. 1700016, 2017.

\bibitem{shorthose2022design}
O.~Shorthose, A.~Albini, L.~He, and P.~Maiolino, ``Design of a 3d-printed soft
  robotic hand with integrated distributed tactile sensing,'' \emph{IEEE
  Robotics and Automation Letters}, vol.~7, no.~2, pp. 3945--3952, 2022.

\bibitem{becker2022active}
K.~Becker, C.~Teeple, N.~Charles, Y.~Jung, D.~Baum, J.~C. Weaver, L.~Mahadevan,
  and R.~Wood, ``Active entanglement enables stochastic, topological
  grasping,'' \emph{Proceedings of the National Academy of Sciences}, vol. 119,
  no.~42, p. e2209819119, 2022.

\bibitem{brown2010universal}
E.~Brown, N.~Rodenberg, J.~Amend, A.~Mozeika, E.~Steltz, M.~R. Zakin,
  H.~Lipson, and H.~M. Jaeger, ``Universal robotic gripper based on the jamming
  of granular material,'' \emph{Proceedings of the National Academy of
  Sciences}, vol. 107, no.~44, pp. 18\,809--18\,814, 2010.

\bibitem{ng2023untethered}
C.~S.~X. Ng and G.~Z. Lum, ``Untethered soft robots for future planetary
  explorations?'' \emph{Advanced Intelligent Systems}, vol.~5, no.~3, p.
  2100106, 2023.

\bibitem{zhang2023progress}
Y.~Zhang, P.~Li, J.~Quan, L.~Li, G.~Zhang, and D.~Zhou, ``Progress, challenges,
  and prospects of soft robotics for space applications,'' \emph{Advanced
  Intelligent Systems}, vol.~5, no.~3, p. 2200071, 2023.

\bibitem{toward_perceptive_soft_robots_wang}
H.~Wang, M.~Totaro, and L.~Beccai, ``Toward perceptive soft robots: Progress
  and challenges,'' \emph{Advanced Science}, vol.~5, no.~9, p. 1800541, 2018.

\bibitem{flexible_sensing_qu}
\BIBentryALTinterwordspacing
J.~Qu, G.~Cui, Z.~Li, S.~Fang, X.~Zhang, A.~Liu, M.~Han, H.~Liu, X.~Wang, and
  X.~Wang, ``Advanced flexible sensing technologies for soft robots,''
  \emph{Advanced Functional Materials}, vol. n/a, no. n/a, p. 2401311, 2024.
  [Online]. Available:
  \url{https://onlinelibrary.wiley.com/doi/abs/10.1002/adfm.202401311}
\BIBentrySTDinterwordspacing

\bibitem{Truby2019}
R.~L. Truby, R.~K. Katzschmann, J.~A. Lewis, and D.~Rus, ``Soft robotic fingers
  with embedded ionogel sensors and discrete actuation modes for
  somatosensitive manipulation,'' in \emph{2019 2nd IEEE International
  Conference on Soft Robotics (RoboSoft)}.\hskip 1em plus 0.5em minus
  0.4em\relax IEEE, 4 2019, pp. 322--329.

\bibitem{Toshimitsu2021}
Y.~Toshimitsu, K.~W. Wong, T.~Buchner, and R.~Katzschmann, ``Sopra: Fabrication
  \& dynamical modeling of a scalable soft continuum robotic arm with
  integrated proprioceptive sensing,'' in \emph{2021 IEEE/RSJ International
  Conference on Intelligent Robots and Systems (IROS)}.\hskip 1em plus 0.5em
  minus 0.4em\relax IEEE, 9 2021, pp. 653--660.

\bibitem{Georgopoulou2023}
A.~Georgopoulou, S.~Hamelryckx, K.~Junge, L.~M. Eckey, S.~Rogler,
  R.~Katzschmann, J.~Hughes, and F.~Clemens, ``A multi-material robotic finger
  with integrated proprioceptive and tactile capabilities produced with a
  circular process,'' in \emph{2023 IEEE International Conference on Soft
  Robotics (RoboSoft)}.\hskip 1em plus 0.5em minus 0.4em\relax IEEE, 4 2023,
  pp. 1--6.

\bibitem{optoelectronically_innervated_hand_zhao}
H.~Zhao, K.~O’Brien, S.~Li, and R.~F. Shepherd, ``Optoelectronically
  innervated soft prosthetic hand via stretchable optical waveguides,''
  \emph{Science Robotics}, vol.~1, no.~1, p. eaai7529, 2016.

\bibitem{Hegde2023}
C.~Hegde, J.~Su, J.~M.~R. Tan, K.~He, X.~Chen, and S.~Magdassi, ``Sensing in
  soft robotics,'' \emph{ACS Nano}, vol.~17, pp. 15\,277--15\,307, 8 2023.

\bibitem{macrobend_optical_sensing_sareh}
S.~Sareh, Y.~Noh, M.~Li, T.~Ranzani, H.~Liu, and K.~Althoefer, ``Macrobend
  optical sensing for pose measurement in soft robot arms,'' \emph{Smart
  Materials and Structures}, vol.~24, p. 125024, 11 2015.

\bibitem{Yang2020}
Z.~Yang, S.~Ge, F.~Wan, Y.~Liu, and C.~Song, ``Scalable tactile sensing for an
  omni-adaptive soft robot finger,'' in \emph{2020 3rd IEEE International
  Conference on Soft Robotics (RoboSoft)}.\hskip 1em plus 0.5em minus
  0.4em\relax IEEE, 5 2020, pp. 572--577.

\bibitem{Rich2018}
S.~I. Rich, R.~J. Wood, and C.~Majidi, ``Untethered soft robotics,''
  \emph{Nature Electronics}, vol.~1, pp. 102--112, 2 2018.

\bibitem{Yun2021}
S.~Yun, J.~Jeong, S.~Mun, and K.-U. Kyung, ``A highly stretchable optical
  strain sensor monitoring dynamically large strain for
  deformation-controllable soft actuator,'' \emph{Smart Materials and
  Structures}, vol.~30, p. 105020, 10 2021.

\bibitem{Jung2020}
J.~Jung, M.~Park, D.~Kim, and Y.-L. Park, ``Optically sensorized elastomer air
  chamber for proprioceptive sensing of soft pneumatic actuators,'' \emph{IEEE
  Robotics and Automation Letters}, vol.~5, pp. 2333--2340, 4 2020.

\bibitem{DelBono2024}
V.~D. Bono, M.~McCandless, F.~J. Wise, and S.~Russo, ``A soft miniaturized
  continuum robot with 3d shape sensing via functionalized soft optical
  waveguides,'' in \emph{2024 IEEE International Conference on Robotics and
  Automation (ICRA)}.\hskip 1em plus 0.5em minus 0.4em\relax IEEE, 5 2024, pp.
  5309--5316.

\bibitem{Kang2023}
J.~Kang, S.~Lee, and Y.-L. Park, ``Soft bending actuator with fiber-jamming
  variable stiffness and fiber-optic proprioception,'' \emph{IEEE Robotics and
  Automation Letters}, vol.~8, pp. 7344--7351, 11 2023.

\bibitem{Yi2023}
J.~Yi, B.~Kim, K.~J. Cho, and Y.~L. Park, ``Underactuated robotic gripper with
  fiber-optic force sensing tendons,'' \emph{IEEE Robotics and Automation
  Letters}, vol.~8, pp. 7607--7614, 11 2023.

\bibitem{Han2024}
M.~S. Han, J.~T. Lin, and C.~K. Harnett, ``A bio-inspired robotic finger driven
  and shape-sensed by soft optical tendons,'' \emph{2024 IEEE 7th International
  Conference on Soft Robotics, RoboSoft 2024}, pp. 165--170, 2024.

\bibitem{Lipson2015}
H.~Lipson, ``A grand challenge for additive manufacturing: Can we print a robot
  that will walk out of the printer?'' \emph{3D Printing and Additive
  Manufacturing}, vol.~2, pp. 41--41, 6 2015.

\bibitem{Rus2015}
D.~Rus and M.~T. Tolley, ``Design, fabrication and control of soft robots,''
  \emph{Nature}, vol. 521, pp. 467--475, 5 2015.

\bibitem{Muth2014}
J.~T. Muth, D.~M. Vogt, R.~L. Truby, Y.~Mengüç, D.~B. Kolesky, R.~J. Wood,
  and J.~A. Lewis, ``Embedded 3d printing of strain sensors within highly
  stretchable elastomers,'' \emph{Advanced Materials}, vol.~26, pp. 6307--6312,
  9 2014.

\bibitem{Shih2019}
B.~Shih, C.~Christianson, K.~Gillespie, S.~Lee, J.~Mayeda, Z.~Huo, and M.~T.
  Tolley, ``Design considerations for 3d printed, soft, multimaterial resistive
  sensors for soft robotics,'' \emph{Frontiers in Robotics and AI}, vol.~6, 4
  2019.

\bibitem{Ntagios2020}
M.~Ntagios, H.~Nassar, A.~Pullanchiyodan, W.~T. Navaraj, and R.~Dahiya,
  ``Robotic hands with intrinsic tactile sensing via 3d printed soft pressure
  sensors,'' \emph{Advanced Intelligent Systems}, vol.~2, p. 1900080, 6 2020.

\bibitem{Woodman2024}
S.~J. Woodman, D.~S. Shah, M.~Landesberg, A.~Agrawala, and R.~Kramer-Bottiglio,
  ``Stretchable arduinos embedded in soft robots,'' \emph{Science Robotics},
  vol.~9, 9 2024.

\bibitem{Soter2018}
G.~Soter, A.~Conn, H.~Hauser, and J.~Rossiter, ``Bodily aware soft robots:
  Integration of proprioceptive and exteroceptive sensors,'' in \emph{2018 IEEE
  International Conference on Robotics and Automation (ICRA)}.\hskip 1em plus
  0.5em minus 0.4em\relax IEEE, 5 2018, pp. 2448--2453.

\bibitem{Thuruthel2019}
T.~G. Thuruthel, B.~Shih, C.~Laschi, and M.~T. Tolley, ``Soft robot perception
  using embedded soft sensors and recurrent neural networks,'' \emph{Science
  Robotics}, vol.~4, 1 2019.

\bibitem{Truby2020}
R.~L. Truby, C.~D. Santina, and D.~Rus, ``Distributed proprioception of 3d
  configuration in soft, sensorized robots via deep learning,'' \emph{IEEE
  Robotics and Automation Letters}, vol.~5, pp. 3299--3306, 4 2020.

\bibitem{ArUco}
S.~Garrido-Jurado, R.~Muñoz-Salinas, F.~Madrid-Cuevas, and M.~Marín-Jiménez,
  ``Automatic generation and detection of highly reliable fiducial markers
  under occlusion,'' \emph{Pattern Recognition}, vol.~47, pp. 2280--2292, 6
  2014.

\bibitem{Li2023}
W.~Li, D.~Hu, and L.~Yang, ``Actuation mechanisms and applications for soft
  robots: A comprehensive review,'' \emph{Applied Sciences}, vol.~13, p. 9255,
  8 2023.

\bibitem{Fleming2013}
A.~J. Fleming, ``A review of nanometer resolution position sensors: Operation
  and performance,'' \emph{Sensors and Actuators A: Physical}, vol. 190, pp.
  106--126, 2 2013.

\bibitem{Yin2014}
S.~Yin, S.~X. Ding, X.~Xie, and H.~Luo, ``A review on basic data-driven
  approaches for industrial process monitoring,'' \emph{IEEE Transactions on
  Industrial Electronics}, vol.~61, pp. 6418--6428, 11 2014.

\end{thebibliography}

\end{document}